\documentclass[letterpaper]{article}
\usepackage{graphcert-preprint}
\usepackage[hyphens]{url}
\usepackage{graphicx}
\usepackage{amsmath}
\usepackage{multirow}
\usepackage{dsfont}
\usepackage{array}
\usepackage{natbib}
\usepackage{caption}
\usepackage{algorithm}
\usepackage{algorithmic}
\usepackage{xcolor}
\usepackage{newfloat}
\usepackage{listings}
\DeclareCaptionStyle{ruled}{labelfont=normalfont,labelsep=colon,strut=off}
\floatstyle{ruled}
\newfloat{listing}{tb}{lst}{}
\floatname{listing}{Listing}
\usepackage{booktabs}

\newcommand{\mathbb}{\mathds}

\title{GraphCert: Bootstrap Agentic Graph Reasoning with Certified Evidence Rubrics}

\author{
  Weiqi Jiang\equalcontrib, Yuchen Ying\equalcontrib,
  Rui Wang, Kaixuan Chen, Bingde Hu, Shunyu Liu, Yu Wang,
  and Tongya Zheng\corresponding
}
\affiliations{
  Weiqi Jiang, Yuchen Ying, Rui Wang, Kaixuan Chen, Bingde Hu, and Yu Wang: \\
  Zhejiang University, Hangzhou, China \\
  Shunyu Liu: Nanyang Technological University, Singapore \\
  Tongya Zheng: Hangzhou City University, Hangzhou, China \\
  \texttt{weiqi.jiang@zju.edu.cn, yingyc@zju.edu.cn, chenkx@zju.edu.cn,} \\
  \texttt{shunyu.liu@ntu.edu.sg, yu.wang@zju.edu.cn} \\
  \texttt{rwang21@zju.edu.cn, tonyhu@zju.edu.cn, doujiang\_zheng@163.com}
}

\begin{document}

\maketitle

\begin{abstract}

Graph agents extend large language models (LLMs) with the ability to actively explore and reason over knowledge graphs through multi-step interactions with graph tools.
However, training capable graph agents typically requires large collections of question-answer pairs and reasoning trajectories, whose manual construction is costly and difficult to scale.
Moreover, employing proprietary LLMs to generate such supervision further risks exposing sensitive graph data to external services.
Therefore, we propose GraphCert to bootstrap agentic graph reasoning with certified evidence rubrics during post-training. Specifically, the Bootstrapped Graph Quizzer guided by generation controls produces graph-grounded QA pairs and marks supporting evidence, which undergo execution certification and semantic curation. The accepted evidence is then canonicalized into certified evidence rubrics that later reward Graph Solver evidence alignment alongside answer correctness during GRPO training.
Experiments on five graph reasoning domains in GRBENCH demonstrate that GraphCert consistently outperforms substantially larger LLM agents and post-training method. Furthermore, our analysis demonstrates that the learned policy transfers robustly across heterogeneous graph domains, suggesting that GraphCert acquires reusable graph-reasoning capabilities rather than domain-specific patterns. These results establish executable self-certification as an effective approach to self-training compact graph reasoning agents.
Our code will be made publicly available.
\end{abstract}

\section{Introduction}
Knowledge graphs are widely adopted across domains such as social network analysis~\cite{yang2012youtube}, biomedical discovery~\cite{saha2017protein_motifs}, and recommender systems~\cite{wang2025llm4dsr}. A central challenge in these applications is text-attributed graph reasoning, which seeks to answer complex queries by jointly exploiting graph structure and the rich textual content attached to nodes and edges. This task requires models to extract relevant knowledge from text and integrate it with graph relations for accurate multi-hop reasoning.

Recently, large language models (LLMs)~\cite{achiam2023gpt4,yang2025qwen3} have demonstrated remarkable capabilities in natural language understanding and generation. Despite their impressive performance, their sequence-centric architecture is not naturally aligned with the non-linear and relational structure of graphs. Consequently, enabling LLMs to effectively capture graph topology and perform multi-hop reasoning remains a fundamental challenge. Existing approaches to LLM-based graph reasoning can be broadly categorized into three paradigms: linearizing graph structures into text with prompt-based augmentation~\cite{fatemi2024talk}, querying the underlying graph through specialized tools~\cite{jin2024graphcot}, and generating general-purpose code to manipulate graph data~\cite{finkelshtein2026graphascode}. Despite recent progress, these approaches remain susceptible to structural information loss~\cite{guo2023gpt4graph}, cascading errors in multi-hop reasoning, and failures arising from insufficient awareness of graph schema information, such as edge semantics. More importantly, their effectiveness often depends on external representations, tools, and programs, rather than intrinsic and transferable graph reasoning capabilities. This motivates dedicated post-training to equip LLMs with intrinsic abilities for graph reasoning.

Accordingly, recent studies~\cite{bai2026graphdancer,liu2025grapho1} have explored dedicated post-training strategies for graph reasoning, including supervised fine-tuning, PPO, and GRPO. However, these approaches build upon manually designed question--answer pairs from established benchmarks, whose limited scale and question patterns provide insufficient coverage of heterogeneous graph structures and compositional reasoning requirements. Automatically generating graph-reasoning supervision offers a promising way to alleviate this bottleneck~\cite{lu2025deepdive,graphscout2026}. Although such approaches alleviate data scarcity, limited question diversity, and manual annotation costs, their reliance on powerful external LLMs presents substantial barriers to real-world deployment.

These practical constraints motivate shifting supervision generation from powerful external teachers to compact, locally deployable models. However, the supervision generated by such models can be noisy. As shown in our experiments, certification rejects over one third of the Quizzer-generated candidates in both Healthcare and Literature. Of all candidates, 21.2\% in Healthcare and 14.4\% in Literature are rejected because of hard failures, including schema-direction errors, answer--evidence inconsistencies, incomplete evidence, and execution failures. Beyond these hard failures, additional candidates are filtered out during semantic curation. Without effective verification and curation, such noise may be amplified during self-training. This leads to the central question of this work: can a compact local model generate diverse graph-reasoning supervision, reliably certify its correctness, and learn from it without relying on an external teacher?

To address this challenge, we propose \textbf{GraphCert}, a self-training framework that bootstraps agentic graph reasoning with certified evidence rubrics, thereby enabling the training of a compact, locally deployable LLM without external teacher models or human-authored reasoning trajectories. GraphCert treats self-generated supervision as candidate certificates rather than trusted labels and structures self-training around three components. The \textbf{Bootstrapped Graph Quizzer} explores the graph under sampled generation rubrics, produces graph-grounded question--answer pairs, and marks the evidence supporting each answer. The resulting candidate certificates undergo execution certification and semantic curation, after which accepted evidence is canonicalized into \textbf{Certified Evidence Rubrics}. The \textbf{Graph Solver with Recorded Evidence} then independently explores the graph, predicts answers, and marks the evidence supporting its predictions. The \textbf{Certified Rubric-Augmented Reward} compares the Solver's recorded evidence against the hidden certified rubrics and augments answer correctness with executable-rubric alignment during GRPO optimization. Thus, certified evidence serves both as a quality-control mechanism for self-generated supervision and as structural guidance for Solver training. Experiments on GRBENCH show that GraphCert substantially improves the compact base model, outperforms larger or teacher-dependent graph agents, and transfers robustly across heterogeneous graph domains.
In summary, our contributions are as follows:
\begin{itemize}

    \item We identify a fundamental limitation of existing training approaches for graph reasoning: they often rely on human-authored supervision or powerful proprietary LLMs to generate training data, which introduces substantial annotation cost, deployment overhead, and privacy concerns.

    \item We propose \textbf{GraphCert}, an executable self-training framework built around certified evidence rubrics that enables compact LLMs to generate, verify, and learn from graph-grounded supervision without external teachers or human-authored reasoning trajectories.

    \item Experiments on GRBENCH show that GraphCert substantially improves compact LLMs, achieves competitive performance against larger or teacher-dependent baselines, and achieves an average relative gain of \textbf{8.2\%} in F1 metric over the strongest baseline on each domain and robust cross-domain transfer.
    
\end{itemize}

\begin{figure*}[t]
    \centering
    \includegraphics[width=1.0\textwidth]{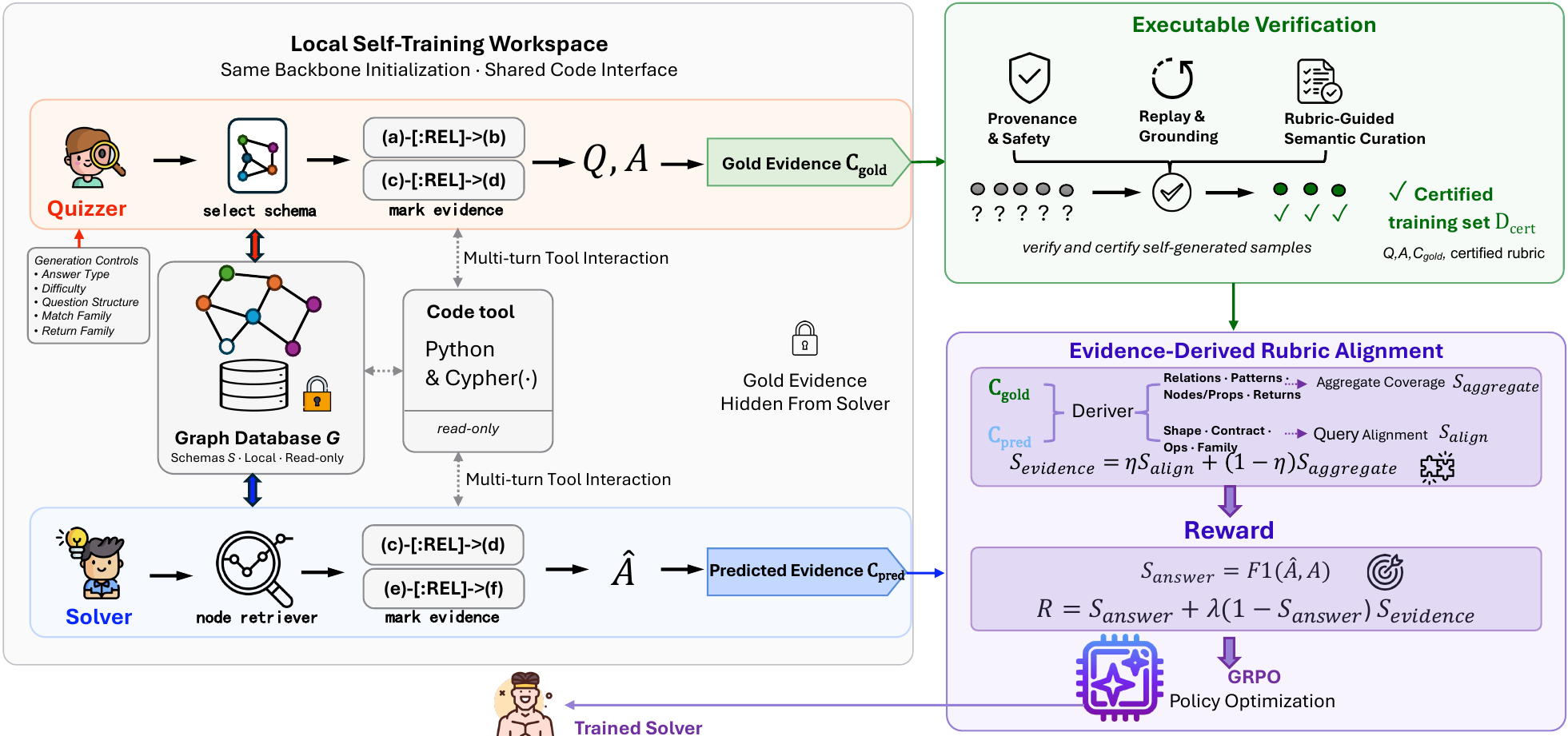}
    \caption{Overview of GraphCert. A local Quizzer generates graph-grounded QA candidates and supporting evidence, which are certified and converted into evidence rubrics. These rubrics provide hidden structural supervision for evidence-aware Solver post-training.}
    \label{fig:framework}
\end{figure*}

\section{Related Work}
\paragraph{LLMs for Graph Reasoning}
Graphs provide a universal abstraction for relational data and are widely used to represent structured knowledge across domains.
With the recent success of large language models (LLMs)~\cite{achiam2023gpt4,yang2025qwen3}, growing attention has been devoted to applying them to graph-related tasks.
Existing studies can be broadly categorized according to the role LLMs play in graph reasoning. One line of work uses LLMs as feature extractors~\cite{duan2023llm_extractor4,chen2024llm_extractor1,ying2026neural}, encoding textual node or edge information before feeding it into GNNs~\cite{kipf2016gcn, xu2018gin}. Such cascaded architectures combine the semantic representations of LLMs with the structural modeling capabilities of GNNs.
Another line of work employs LLMs as direct graph reasoners~\cite{wang2023can, dai2024large}. Some methods serialize graph structures into textual sequences and process them directly with LLMs~\cite{luo2024graphinstruct, huang2025graphthought,guo2025g1}, while others retain graphs in external data structures or graph databases and enable LLMs to access and reason over them through tool calls~\cite{jin2024graphcot, liu2025polyg}. These approaches have been applied to tasks such as node classification~\cite{finkelshtein2026graphascode}, link prediction~\cite{he2024linkgpt}, graph algorithm execution~\cite{zhang2024gcoder, li2025graphteam}, and knowledge graph reasoning~\cite{gao2025graphCounselor,ding2026code}.

\paragraph{Post-Training for Agentic Graph Reasoning}
Despite the growing use of LLMs for graph reasoning, general-purpose models often remain limited in structural exploration and multi-step interaction, motivating task-specific post-training. Some studies train LLMs on synthetic and verifiable graph algorithm problems~\cite{guo2025g1}, while agentic GraphRAG methods optimize iterative graph retrieval and answer generation through supervised fine-tuning or reinforcement learning~\cite{luo2025graphr1,yu2026graphragr1,park2026prograph}. More closely related to our setting, recent methods train knowledge graph agents using annotated paths, synthetic trajectories, or outcome- and process-level rewards~\cite{luo2024rog, liu2025grapho1,bai2026graphdancer}. However, these approaches typically depend on predefined paths, teacher-generated trajectories, or task-specific QA supervision. GraphScout~\cite{graphscout2026} autonomously generates graph-grounded question--answer pairs, but its reliance on proprietary LLMs may raise privacy concerns and inference costs. In contrast, our framework enables compact open-source LLMs to generate executable supervision and improve through self-training, offering a more private and cost-effective solution.

\section{Methodology}

As illustrated in Figure~\ref{fig:framework}, GraphCert comprises a Bootstrapped Graph Quizzer, Executable Verification, and evidence-derived Solver training. The Quizzer generates QA candidates with supporting evidence; Executable Verification certifies them through replay, grounding, and semantic curation; and evidence-derived rubric alignment augments answer correctness during GRPO. Generation rubrics softly guide Quizzer exploration, whereas certified evidence rubrics evaluate Solver evidence.

\subsection{Problem Formulation}

We study question answering over a labeled property graph
\begin{equation}
\mathcal{G}=(\mathcal{V},\mathcal{E},\mathcal{S}),
\label{eq:graph}
\end{equation}
where $\mathcal{V}$ and $\mathcal{E}$ are the node and directed-edge sets, and $\mathcal{S}$ specifies node labels, relation types, properties, directions, and endpoint constraints. Given a natural-language question $Q$, a Solver policy $\pi_\theta$ interacts with $\mathcal{G}$ and produces
\begin{equation}
\tau=(Q,a_1,o_1,\ldots,a_T,o_T,\hat{A}),
\label{eq:trajectory}
\end{equation}
where $a_t$ and $o_t$ are the action and observation at turn $t$, $\hat{A}$ is the Solver answer, and $A$ denotes the execution-grounded reference answer.

\subsection{Bootstrapped Graph Quizzer}
The Bootstrapped Graph Quizzer addresses the first challenge of self-training: how to explore a large graph and construct diverse graph-grounded supervision with a compact local model. Guided by generation controls, the Quizzer explores the graph, produces a candidate question--answer pair $(Q,A)$, and records the executed Cypher results that support the answer. Its output is treated as a candidate certificate rather than directly trusted training data.

\textbf{Self-quizzing settings.} To encourage diverse graph exploration, the Quizzer first samples a generation self-quizzing setting specifying the desired answer type, difficulty, question structure, match family, and return family. Following GraphScout~\cite{graphscout2026}, we consider $\langle h,\_,\_\rangle$, $\langle h,r,*\rangle$, $\langle h,*,t\rangle$, $\langle h,r,t\rangle$, and hybrid structural patterns. We further introduce CypherBench-inspired~\cite{feng-etal-2025-cypherbench} \emph{match-family} and \emph{return-family} controls. Match families characterize graph access, including one-hop relations, multi-hop paths, intersections, comparisons, and optional matches; return families characterize outputs such as properties, distinct sets, aggregations, rankings, and Boolean decisions. These targets diversify generation but remain soft rubric-guided controls rather than trusted labels. 

\textbf{Quizzer interaction tools.} Given a random seed node and its 1-hop neighborhood, the Quizzer uses $\operatorname{select\_schema}$ to inspect detailed information about relevant properties, relation directions, and endpoint constraints. The $\operatorname{code}$ tool provides persistent Python with a synchronous read-only $\operatorname{cypher}(\cdot)$ function, enabling multi-turn querying, result composition, and $\operatorname{mark\_evidence}(\cdot)$ calls on results supporting the answer. Marked results are retained for certification and reward construction.

\textbf{Quizzer certificate.} The Quizzer outputs $(Q,A)$ and marked evidence $\mathcal{C}_{\mathrm{gold}}$. Each trace stores the query, parameters, return metadata, execution status, and provenance, preventing fabricated evidence. The execution-grounded answer must match $A$, and the resulting certificate is verified and semantically curated before training.

\subsection{Certified Evidence Rubrics}
Because the Quizzer is a compact self-generating model, its candidate certificates may contain invalid queries, unsupported answers, or semantically mismatched evidence. We therefore certify each candidate at both the execution and semantic levels, and convert the accepted evidence into a canonical rubric for subsequent Solver supervision. Candidates that pass both procedures are retained in the certified training set $\mathcal{D}_{\mathrm{cert}}$.

\textbf{Execution certification.}
Deterministic checks validate certificate format, trace provenance, schema usage, and read-only query safety. Each marked query is then replayed against $\mathcal{G}$. A certificate receives \emph{judgeable strict} status if replay reproduces the recorded result, the execution-grounded answer matches $A$, and no truncated preview is used as a complete set or exact count. If execution and answer grounding succeed but complex Python composition cannot be fully established statically, the certificate is assigned \emph{judgeable partial} status. Syntax, safety, replay, or answer-grounding failures are treated as hard failures.

\textbf{Semantic curation.}
Both \emph{judgeable strict} and \emph{judgeable partial} candidates that pass execution verification are reviewed by a rubric-guided LLM-as-judge using the same local model as the Quizzer. The Judge is isolated from the full exploration trajectory to reduce bias introduced by intermediate reasoning. It then assesses whether the marked evidence semantically supports the question–answer pair $(Q,A)$, according to rubric criteria such as relation-path faithfulness. Only examples that pass this review are retained and labeled as \emph{accepted strict/partial}.

\textbf{Executable evidence rubrics.}
The shared Cypher deriver converts each marked query and its traced result into an executable evidence rubric by removing superficial differences such as variable names and aliases and extracting graph patterns, return structure, operations, match/return features, and the return contract. Accepted examples form $\mathcal{D}_{\mathrm{cert}}$, retaining $(Q,A)$, gold evidence, evidence-derived query structures, and verifier status. The question--answer pair provides outcome-level supervision, while the hidden query structures provide process-level supervision for the Solver.

\subsection{Graph Solver with Recorded Evidence}
Given the certified questions produced by the preceding components, the same compact model is post-trained as a Graph Solver. For each question $Q$, the Solver independently explores the graph, predicts an answer $\hat{A}$, and records the executed results that support its prediction, without access to the Quizzer's gold evidence or certified rubric.

\textbf{Solver interaction tools.} To support independent graph exploration, the Solver interacts with the graph through entity retrieval and a read-only code interface. The $\operatorname{node\_retriever}$ maps natural-language entity mentions to candidate graph nodes and returns the top-$k$ identifiers as anchors for subsequent queries. The $\operatorname{code}$ tool provides the same persistent Python environment and read-only $\operatorname{cypher}(\cdot)$ function used by the Quizzer, allowing the Solver to execute and compose graph queries across multiple turns. Once it obtains sufficient support for its prediction, the Solver invokes $\operatorname{mark\_evidence}(\cdot)$ on the corresponding query results.

\textbf{Solver evidence.} The Solver finally returns its predicted answer $\hat{A}$ together with the recorded evidence set $\mathcal{C}_{\mathrm{pred}}$. Only explicitly marked results are retained for supervision, while unmarked exploratory queries are excluded. Because the Quizzer and Solver share the same evidence-marking interface and Cypher deriver, $\mathcal{C}_{\mathrm{pred}}$ can be converted into the same canonical rubric representation as the certified evidence. This enables direct structural comparison with the hidden certified rubric in the following reward design.

\subsection{Certified Rubric-Augmented Reward}
The certified evidence rubrics provide structural supervision for Solver post-training. Specifically, GraphCert compares the rubric derived from the Solver's recorded evidence with the hidden certified rubric, and uses their alignment to augment the answer-correctness reward.

\textbf{Executable rubric alignment.}
Let $\mathcal{C}_{\mathrm{gold}}=\{c_i^{\mathrm{gold}}\}_{i=1}^{m}$ and $\mathcal{C}_{\mathrm{pred}}=\{c_j^{\mathrm{pred}}\}_{j=1}^{n}$. To measure whether the Solver recovers the graph structures required by the certified QA pair, we compare every predicted query with every gold query in their canonical structural form. The pair score is a weighted combination of four interpretable comparisons:
\begin{equation}
\begin{aligned}
S_{\mathrm{pair}}(i,j)={}&w_{\mathrm{shape}}S_{\mathrm{shape}}(i,j)
+w_{\mathrm{contract}}S_{\mathrm{contract}}(i,j)\\
&+w_{\mathrm{ops}}S_{\mathrm{ops}}(i,j)
+w_{\mathrm{family}}S_{\mathrm{family}}(i,j).
\end{aligned}
\label{eq:pair-score}
\end{equation}
The non-negative weights sum to one and remain fixed during training and evaluation. Each component captures a distinct aspect of structural agreement: $S_{\mathrm{shape}}$ compares typed and directed graph patterns and participating nodes; $S_{\mathrm{contract}}$ compares return expressions and output requirements; $S_{\mathrm{ops}}$ compares active query operations; and $S_{\mathrm{family}}$ checks match/return feature families. The shape and operation scores are computed using weighted multiset F1, while the contract score combines multiset F1 over return expressions with exact agreement on output shape, properties, deduplication, and ordering. The family score uses exact family agreement. All components are normalized to $[0,1]$, with invalid derivations assigned zero.

We aggregate the pairwise query scores into a set-level alignment score using maximum-weight one-to-one matching. The \emph{query alignment} score uses maximum-weight one-to-one matching between the two query sets. If $\mathcal{M}^{*}$ is the selected matching, then
\begin{equation}
S_{\mathrm{align}}
=\frac{\sum_{(i,j)\in\mathcal{M}^{*}}S_{\mathrm{pair}}(i,j)}
{\max(m,n,1)}.
\label{eq:alignment-score}
\end{equation}
Normalizing by the larger set size penalizes both missing gold queries and superfluous predicted queries while remaining invariant to query order.

Equivalent reasoning may decompose the same task into different queries. We therefore also compute \emph{aggregate coverage} by pooling each side and comparing its relation types, graph patterns, nodes and properties, and return requirements. Specifically, $S_{\mathrm{aggregate}}$ is a weighted combination of multiset-F1 scores over these pooled canonical features. Because pooling discards individual query boundaries, this term can credit structurally equivalent one-to-many or many-to-one query decompositions even when their query-level alignment is imperfect. With mixture weight $\eta\in[0,1]$, the evidence score combines query-level precision with tolerance for equivalent decompositions:
\begin{equation}
S_{\mathrm{evidence}}
=\eta S_{\mathrm{align}}+(1-\eta)S_{\mathrm{aggregate}}.
\label{eq:evidence-score}
\end{equation}

\textbf{Reward design.}
The answer score is $S_{\mathrm{answer}}=\operatorname{F1}(\hat{A},A)$. Given $S_{\mathrm{answer}},S_{\mathrm{evidence}}\in[0,1]$ and evidence weight $\lambda\in[0,1]$, the rollout reward is
\begin{equation}
R(\tau)=S_{\mathrm{answer}}
+\lambda(1-S_{\mathrm{answer}})S_{\mathrm{evidence}}.
\label{eq:final-reward}
\end{equation}
The answer score provides the primary reward, while evidence alignment contributes only a weighted fraction of the remaining gap to one. Consequently, $R(\tau)\in[0,1]$ without explicit clipping. A fully correct answer receives $R(\tau)=1$ regardless of evidence alignment; otherwise, evidence can provide additional training signal without replacing answer correctness. The weight $\lambda$ bounds the reward obtainable from evidence alone. Detailed component formulas and weight settings are provided in the appendix.

\subsection{GRPO Optimization}

For each question, GRPO samples a group of Solver trajectories and normalizes their rewards within the group. The resulting trajectory-level advantage is broadcast across valid response tokens. Let $\rho_{i,t}(\theta)$ be the new-to-old policy ratio and $\bar{\rho}_{i,t}(\theta)=\operatorname{clip}(\rho_{i,t}(\theta),1-\epsilon,1+\epsilon)$ its clipped value. The core objective is
\begin{equation}
\mathcal{L}_{\mathrm{GRPO}}(\theta)
=-\operatorname{E}_{i,t}\!\left[
\min\!\left(\rho_{i,t}(\theta)\operatorname{Adv}_i,
\bar{\rho}_{i,t}(\theta)\operatorname{Adv}_i\right)\right].
\label{eq:grpo-objective}
\end{equation}
Here $\operatorname{Adv}_i$ is the group-normalized reward of trajectory $\tau_i$, and the expectation averages over sampled trajectories and valid response tokens. We additionally apply standard reference-policy KL regularization. Training on $\mathcal{D}_{\mathrm{cert}}$ completes the self-training pipeline: Quizzer-generated certificates become verified hidden supervision, which improves the Solver for subsequent Solver policy learning.

\section{Experiments}

In this section, we evaluate GraphCert by addressing three research questions. 
\begin{itemize}
    \item \textbf{RQ1: Overall effectiveness.} How effectively does GraphCert perform compared with prompting-based and post-trained baselines?
    
    \item \textbf{RQ2: Certification and reward effectiveness.} How much do executable certification and rubric-guided rewards contribute to the method's performance? 
    \item  \textbf{RQ3: Compact self-training and transfer.} How effectively does GraphCert leverage self-generated supervision from a compact Quizzer, and how well does the learned policy transfer across graph domains?
\end{itemize}

\subsection{Experimental Setup}

\textbf{Datasets and metrics.}
We evaluate on GRBENCH~\cite{jin2024graphcot}, which contains 1,740 English questions at three difficulty levels over five knowledge-graph domains: Healthcare, Literature, Academic, E-Commerce, and Legal. Its ten real-world, text-attributed graphs vary substantially in scale and topology. They range from approximately 47K nodes in Healthcare to 84M nodes in Legal. The questions require graph-grounded path traversal, attribute reasoning, and aggregation across heterogeneous schemas. We report token-level F1 and QwenScore. F1 measures normalized token overlap with the reference answer, while QwenScore is the proportion of answers judged correct by Qwen-Max. 

\textbf{Model and training settings.}
We use Qwen3-4B-Instruct-2507 as both the Quizzer and the initial checkpoint of Solver. The verl~\cite{verl} framework is adopted for its implementation of Group Relative Policy Optimization (GRPO) for reinforcement learning. We train the model for up to 400 steps. Detailed training hyperparameters and configuration are provided in appendix.

\textbf{Baselines.}
We compare with complementary graph-access strategies. BaseLLM uses no graph context. TextRAG~\cite{ragsurveytextrag} and GraphRAG~\cite{ye-etal-2024-language-graphrag} retrieve linearized text or subgraphs, while Cypher and PolyG~\cite{liu2025polyg} generate an executable query. GraphCoT~\cite{jin2024graphcot} and GraphCounselor~\cite{gao2025graphCounselor} support iterative graph agent exploration. GraphScout~\cite{graphscout2026} is the closest training-based baseline. We reproduce GraphScout with Qwen3-4B-Instruct-2507 as the Quizzer and as the Solver initialization, matching the compact-Quizzer setting used by GraphCert. The GPT-4o~\cite{achiam2023gpt4} and DeepSeek-V3.2~\cite{2025deepseekv32} prompting results in Table~\ref{tab:main-results} are quoted from GraphScout, whereas the GraphScout(4B) row is our reproduction. We use the published GraphScout result with a DeepSeek-V3.2 Quizzer in the controlled Quizzer comparison in Table~\ref{tab:quizzer-comparison}.

\begin{table*}[!t]
\centering
\small
\renewcommand{\arraystretch}{1.05}
\setlength{\tabcolsep}{0.8mm}
\begin{tabular}{lccccccccccc}
\toprule
& & \multicolumn{2}{c}{Healthcare} & \multicolumn{2}{c}{Literature} & \multicolumn{2}{c}{Academic} & \multicolumn{2}{c}{E-Commerce} & \multicolumn{2}{c}{Legal} \\
\cmidrule(lr){3-4}\cmidrule(lr){5-6}\cmidrule(lr){7-8}\cmidrule(lr){9-10}\cmidrule(lr){11-12}
Method & LLM & QwenScore & F1 & QwenScore & F1 & QwenScore & F1 & QwenScore & F1 & QwenScore & F1 \\
\midrule
\multirow{2}{*}{BaseLLM} & GPT-4o & 0.137 & 0.048 & 0.221 & 0.064 & 0.097 & 0.080 & 0.110 & 0.080 & 0.244 & 0.110 \\
& DeepSeek-V3.2 & 0.104 & 0.075 & 0.192 & 0.111 & 0.134 & 0.140 & 0.095 & 0.115 & 0.222 & 0.265 \\
\addlinespace[1pt]
\multirow{2}{*}{TextRAG} & GPT-4o & 0.074 & 0.059 & 0.179 & 0.116 & 0.098 & 0.090 & 0.200 & 0.181 & 0.256 & 0.232 \\
& DeepSeek-V3.2 & 0.085 & 0.060 & 0.196 & 0.137 & 0.146 & 0.158 & 0.115 & 0.138 & 0.333 & 0.365 \\
\addlinespace[1pt]
\multirow{2}{*}{GraphRAG} & GPT-4o & 0.156 & 0.129 & 0.217 & 0.136 & 0.105 & 0.092 & 0.315 & 0.308 & 0.239 & 0.233 \\
& DeepSeek-V3.2 & 0.170 & 0.140 & 0.221 & 0.151 & 0.177 & 0.164 & 0.165 & 0.167 & 0.361 & 0.381 \\
\addlinespace[1pt]
\multirow{2}{*}{Cypher} & GPT-4o & 0.311 & 0.107 & 0.533 & 0.157 & 0.605 & 0.197 & 0.485 & 0.101 & 0.389 & 0.118 \\
& DeepSeek-V3.2 & 0.422 & 0.097 & 0.583 & 0.144 & 0.580 & 0.186 & \underline{0.545}& 0.114 & 0.350 & 0.094 \\
\addlinespace[1pt]
\multirow{2}{*}{GraphCoT} & GPT-4o & 0.415 & 0.447 & 0.463 & 0.395 & 0.587 & 0.547 & 0.410 & 0.360 & 0.494 & 0.403 \\
& DeepSeek-V3.2 & 0.441 & 0.443 & 0.563 & 0.466 & 0.628 & 0.546 & 0.470 & 0.418 & 0.555 & 0.559 \\
\addlinespace[1pt]
\multirow{2}{*}{PolyG} & GPT-4o & 0.378 & 0.107 & 0.463 & 0.119 & 0.516 & 0.138 & 0.440 & 0.092 & 0.344 & 0.087 \\
& DeepSeek-V3.2 & 0.493 & 0.104 & 0.533 & 0.110 & 0.585 & 0.146 & 0.445 & 0.087 & 0.339 & 0.070 \\
\addlinespace[1pt]
\multirow{2}{*}{GraphCounselor} & GPT-4o & 0.485 & 0.441 & 0.592 & 0.438 & 0.644 & 0.605 & 0.490 & 0.381 & 0.433 & 0.219 \\
& DeepSeek-V3.2 & 0.470 & 0.452 & 0.621 & 0.578 & \textbf{0.657} & \underline{0.659} & 0.510& \underline{0.489} & 0.483 & 0.286 \\
\shortstack[l]{GraphScout(4B)} & Qwen3-4B & \underline{0.692} & \underline{0.689} & \underline{0.629} & \underline{0.618} & 0.568 & 0.579 & 0.500 & 0.481 & \underline{0.595} & \underline{0.621} \\
\midrule
GraphCert(Untrained) & Qwen3-4B & 0.303 & 0.300 & 0.304 & 0.301 & 0.233 & 0.231 & 0.245 & 0.212 & 0.194 & 0.209 \\
\textbf{GraphCert} & Qwen3-4B & \textbf{0.744} & \textbf{0.722} & \textbf{0.656} & \textbf{0.681} & \underline{0.651} & \textbf{0.674} & \textbf{0.566} & \textbf{0.583} & \textbf{0.636} & \textbf{0.648} \\
\bottomrule
\end{tabular}
\caption{Performance Comparison on GRBENCH Across Baseline Methods and GraphCert. Best results are in bold, and second-best results are underlined.}
\label{tab:main-results}
\end{table*}

\subsection{RQ1: Overall Effectiveness}
To answer RQ1, Table~\ref{tab:main-results} compares GraphCert with prompting-, retrieval-, tool-, and training-based baselines across all five domains, considering both absolute performance and gains over the untrained Solver. GraphCert achieves the best F1 on every domain and the best QwenScore on four. On Academic, its QwenScore is only 0.006 below DeepSeek-V3.2-based GraphCounselor, while its F1 remains the highest. The consistent gains across token-overlap and judge-based evaluation suggest that the improvement is not metric-specific. Relative to GraphCert(Untrained), post-training substantially improves both QwenScore and F1 in every domain, showing stronger graph use and answer generation. GraphCert also consistently outperforms GraphScout(4B) under the matched compact-Quizzer setting, suggesting greater robustness to noisy self-generated supervision. Overall, these results answer RQ1 affirmatively: GraphCert turns the weak 4B base agent into a competitive Solver that surpasses larger prompting-based systems and the matched self-training baseline across heterogeneous graph schemas and scales.

\subsection{RQ2: Certification and Reward Effectiveness}

\begin{table}[h]
\centering
\small
\setlength{\tabcolsep}{3mm}
\begin{tabular}{lcc}
\toprule
Statistic & Healthcare & Literature \\
\midrule
Candidates & 3,126 & 3,096 \\
Accepted strict (\%) & 34.87 & 29.40 \\
Accepted partial (\%) & 29.10 & 35.20 \\
Rejected (\%) & 36.03 & 35.40 \\
Certified examples & 2,000 & 2,000 \\
\bottomrule
\end{tabular}
\caption{Pipeline status before training. Strict and partial examples enter $\mathcal{D}_{\mathrm{cert}}$ only after semantic curation. Rejected examples combine program and semantic failures.}
\label{tab:certificate-yield}
\end{table}
To answer RQ2, we evaluate the two core mechanisms of GraphCert from both the data and optimization perspectives. We first examine how certification filters and improves Quizzer-generated supervision, and then use controlled ablations to determine whether certification and certified rubric-based rewards improve downstream Solver performance.

\textbf{Certificate analysis.}
We first measure how executable certification changes the self-generated candidate pool before training.
Table~\ref{tab:certificate-yield} shows that certification rejects 36.03\% of Healthcare candidates and 35.40\% of Literature candidates. The gate therefore removes more than one third of the raw Quizzer output in both domains. These counts establish that certification materially changes the training pool, while the following ablation tests whether this filtering improves the Solver. 
We further inspect the hard failures detected during executable certification. Certification identifies 662 of 3,126 Healthcare candidates (21.2\%) and 446 of 3,096 Literature candidates (14.4\%). The category counts are 8/12 for execution failures, 348/66 for schema-direction errors, 118/314 for answer-evidence inconsistencies, and 188/54 for incomplete evidence in Healthcare/Literature. Schema errors dominate in Healthcare, whereas answer-evidence inconsistencies are most common in Literature. These results highlight the substantial noise in supervision generated by a compact Quizzer.

\begin{table}[h]
\centering
\small
\setlength{\tabcolsep}{0.8mm}
\renewcommand{\arraystretch}{1.05}
\begin{tabular}{lccccc}
\toprule
& $N$ & \shortstack{Answer\\Supported}
& \shortstack{Evidence\\Sufficient}
& \shortstack{Query\\Consistent}
& \shortstack{Strict\\Grounding} \\
\midrule
Plain& 300 & 38.7 & 38.7 & 45.3 & 36.7 \\
GraphCert
& 300 & \textbf{82.7} & \textbf{72.7}
& \textbf{77.3} & \textbf{71.0} \\
\bottomrule
\end{tabular}
\caption{Independent assessment of generated QA quality (\%). 
The blinded LLM judge is used for relative comparison rather than as a calibrated ground-truth oracle. Plain denotes the quizzer without certification or semantic curation.}
\label{tab:qa-quality}
\end{table}

\textbf{Independent QA quality assessment.}
We compare 300 QA pairs generated by a plain Qwen3-4B quizzer (no certification and semantic curation) with 300 examples retained by GraphCert. A blinded DeepSeek-V4-Flash~\cite{deepseekai2026deepseekv4} judge evaluates whether the execution trace supports the answer, provides sufficient evidence, and is consistent with the question, using the same prompt for both methods. As shown in Table~\ref{tab:qa-quality}, GraphCert improves all dimensions by 32.0--44.0 percentage points and nearly doubles the strict grounding rate from 36.7\% to 71.0\%. The gains in evidence sufficiency and query consistency indicate that GraphCert improves the alignment among questions, answers, and graph evidence, rather than merely removing unexecutable queries. Although the conservative judge may reject some valid examples with implicit entity mappings or multi-query composition, the same criteria are applied to both methods, supporting the reliability of the large relative improvement.

\begin{table}[h]
\centering
\small
\setlength{\tabcolsep}{1.8mm}
\begin{tabular}{lcc}
\toprule
Variant & Healthcare & Literature \\
\midrule
GraphCert & \textbf{0.722} & \textbf{0.681} \\
w/o Certification & 0.612 & 0.624 \\
w/o $S_{\mathrm{evidence}}$ & 0.566 & 0.654 \\
\bottomrule
\end{tabular}
\caption{F1 results of controlled ablation experiments on the Healthcare and Literature domains. All variants use the same training data size and optimization budget.}
\label{tab:ablation}
\end{table}

\textbf{Ablation study on certification and rubric-based reward.}
Table~\ref{tab:ablation} compares GraphCert with two controlled variants. In the \emph{w/o Certification} variant, we remove the rubric-guided question-family instructions, verification, and semantic curation, and train the model directly on unfiltered Quizzer outputs. In the \emph{w/o $S_{\mathrm{evidence}}$} variant, we disable the evidence-based reward while keeping the training data unchanged. 
Training on raw Quizzer outputs leads to a clear degradation across both domains, indicating that controlling and verifying self-generated supervision is important before it is used for learning. Removing the evidence reward causes a loss, particularly on Healthcare, suggesting that answer correctness alone does not provide sufficient guidance for learning reliable graph reasoning. Together, these results show that certification and evidence-reward optimization play important roles in GraphCert.

Overall, the results answer RQ2 affirmatively. Certification removes a substantial portion of noisy Quizzer outputs and markedly improves the grounding quality of the retained QA pairs, while the controlled ablations show that both certified training data and rubric-based evidence supervision contribute to Solver performance. GraphCert therefore benefits from complementary improvements on the data-construction and post-training sides.

\subsection{RQ3: Compact Self-Training and Transfer}
To answer RQ3, we examine whether GraphCert can extract effective supervision from a compact self-Quizzer and whether the resulting reasoning policy generalizes beyond its training graph. We first compare compact- and strong-Quizzer settings, and then evaluate cross-domain transfer across all five GRBENCH domains.
\begin{table}[h]
\centering
\small
\renewcommand{\arraystretch}{1.08}
\setlength{\tabcolsep}{1.8mm}
\begin{tabular}{lcc}
\toprule
Method & Literature & E-Commerce \\
\midrule
GraphCert(Untrained) & 0.301 & 0.212 \\
GraphCert & \textbf{0.681} & \textbf{0.583} \\
GraphScout(DS) & 0.646 & 0.562 \\
GraphScout(4B) & 0.618 & 0.481 \\
\bottomrule
\end{tabular}
\caption{Quizzer-scale comparison reported in terms of F1 metric. GraphScout(DS) denotes the original GraphScout which uses DeepSeek-V3.2 as the Quizzer}
\label{tab:quizzer-comparison}
\end{table}

\textbf{Quizzer-scale comparison.}
To examine how effectively GraphCert can learn from supervision generated by a compact Quizzer, Table~\ref{tab:quizzer-comparison} compares GraphCert with GraphScout under both strong- and compact-Quizzer settings. We report Literature and E-Commerce domains. Replacing DeepSeek-V3.2 with Qwen3-4B as the GraphScout Quizzer consistently degrades performance, indicating that its self-training pipeline is sensitive to the capability of the model generating supervision. In contrast, under the same Qwen3-4B Quizzer, GraphCert outperforms GraphScout(4B) and even exceeds the strong-Quizzer GraphScout variant on both domains. This comparison suggests that GraphCert is better able to extract useful training signals from self-generated data produced by a compact model. 

\begin{table}[h]
\centering
\small
\setlength{\tabcolsep}{1.3mm}
\begin{tabular}{lccccc}
\toprule
& \multicolumn{5}{c}{Test Domain} \\
\cmidrule(lr){2-6}
Training Domain & Health. & Lit. & Acad. & E-Com. & Legal \\
\midrule
Healthcare & 0.722 & 0.544 & 0.500 & 0.496 & 0.637 \\
Literature & 0.672 & 0.681 & 0.537 & 0.602 & 0.644 \\
Academic & 0.561 & 0.587 & 0.674 & 0.571 & 0.655 \\
E-Commerce & 0.614 & 0.656 & 0.488 & 0.583 & 0.624 \\
Legal & 0.643 & 0.691 & 0.648 & 0.571 & 0.648 \\
\bottomrule
\end{tabular}
\caption{Cross-domain transfer in terms of F1 metric. Each row specifies the training domain and each column specifies the test domain.}
\label{tab:domain-transfer}
\end{table}

\textbf{Cross-domain transfer.}
We next evaluate each domain-trained Solver on all five GRBENCH domains. Table~\ref{tab:domain-transfer} shows that the learned policy transfers consistently beyond the graph on which it is trained. Strong off-domain performance appears across different training domains rather than being confined to a few favorable source--target pairs, suggesting that the Solver acquires reusable graph-reasoning behaviors that are not tied to a particular schema. At the same time, transfer performance varies across domain pairs, indicating that domain-specific structure still affects how well these behaviors generalize. Overall, the results support cross-domain transfer rather than memorization of a single training graph.

\begin{table}[h]
\centering
\small
\setlength{\tabcolsep}{3mm}
\begin{tabular}{lcc}
\toprule
Backbone & Healthcare & Literature \\
\midrule
Qwen3-4B-Instruct-2507 & 0.744/0.722 & 0.656/0.681 \\
Qwen3-8B & 0.609/0.601 & 0.663/0.687 \\
\bottomrule
\end{tabular}
\caption{Scaling results reported as QwenScore/F1. Both rows use the full certified self-training method.}
\label{tab:model-scaling}
\end{table}

\textbf{Model scaling.}
We evaluate the full method with Qwen3-8B on Healthcare and Literature. As shown in Table~\ref{tab:model-scaling}, Qwen3-8B yields a small improvement on Literature but degrades on Healthcare. A similar post-training scaling pattern has been reported by GraphScout, where Qwen3-8B often underperforms Qwen3-4B-Instruct-2507 after training, and the authors hypothesize that this behavior may be related to the hybrid reasoning configuration of Qwen3-8B~\cite{graphscout2026}.

Together, these results answer RQ3 affirmatively. GraphCert learns stronger policies from compact self-generated supervision than teacher-sensitive self-training alternatives do, while transferring consistently across heterogeneous graph domains. The gains thus reflect reusable graph-exploration and reasoning capabilities rather than reliance on a powerful Quizzer or a single training schema. Supplementary experimental analyses are provided in the appendix.
\section{Conclusion}
This paper presents GraphCert, a self-training framework that enables compact LLMs to generate graph-grounded QA pairs, certify their supporting evidence, and derive evidence rubrics for Solver supervision. Experiments on five GRBENCH domains show that GraphCert consistently outperforms strong prompting-based and self-training baselines, while ablation and transfer studies confirm the benefits of certification, rubric-augmented reward, and cross-domain generalization. These results demonstrate that GraphCert provides an effective approach to training compact graph agents without external teacher models.

\bibliography{references}

\appendix
\section{Quizzer Exploration Settings}

The Quizzer uses a structured generation-control space to encourage diverse graph-grounded questions. The controls specify the desired answer form, abstract query pattern, reasoning difficulty, graph-access family, and return family. These controls guide question generation but are not treated as verified labels. The actual question, answer, and marked evidence must still pass execution certification and semantic curation before entering $\mathcal{D}_{\mathrm{cert}}$.

\textbf{Quizzer sampling.} During graph exploration and initial candidate question--answer generation, we sample the Quizzer with temperature $0.7$ to encourage diversity. Evidence-code retries and evidence-grounded QA regeneration performed while completing the candidate certificate are ordinary agent self-correction, not Quizzer repair.

\textbf{Test-set leakage prevention.} To prevent test-set leakage, we discard a generated candidate if any entity grounded in its question, answer, or marked evidence appears in an official test question.

\textbf{Generation-control space.} Following GraphScout~\cite{graphscout2026}, the Quizzer uses answer type, query pattern, and difficulty as complementary task axes rather than rigid templates. We further add CypherBench-inspired~\cite{feng-etal-2025-cypherbench} match and return families. Table~\ref{tab:quizzer-generation-controls} gives the complete inventories and compatibility mappings. In the pattern notation, $h$, $r$, and $t$ denote the head entity, relation, and tail entity; ``$\_$'' is unspecified, while ``$*$'' must be discovered through graph exploration.

The five controls describe complementary layers of a generated task. Answer type specifies the final answer form, while query pattern specifies which abstract head, relation, and tail slots are given or must be discovered. Match family then describes how evidence is accessed or composed over the graph, and return family describes the structure in which query evidence is returned. Difficulty further restricts the admissible match and return families rather than serving as a verified label of the finished question. Accordingly, answer type constrains return family, query pattern constrains match family, and difficulty constrains both.

\begin{table*}[t]
\centering
\small
\setlength{\tabcolsep}{2.5pt}
\begin{tabular}{@{}>{\raggedright\arraybackslash}p{0.22\textwidth}>{\raggedright\arraybackslash}p{0.755\textwidth}@{}}
\toprule
Control or condition & Values or admissible families \\
\midrule
\multicolumn{2}{@{}l}{\textbf{Base generation axes}} \\
Answer type & \texttt{entity}, \texttt{boolean}, \texttt{number}, \texttt{set} \\
Query pattern & $\langle h,\_,\_\rangle$, $\langle h,r,*\rangle$, $\langle h,*,t\rangle$, $\langle h,r,t\rangle$, \texttt{Hybrid} \\
Difficulty & \texttt{simple}, \texttt{medium}, \texttt{hard} \\
\midrule
\multicolumn{2}{@{}l}{\textbf{Family inventories}} \\
Match family & \texttt{entity\_lookup}, \texttt{one\_hop\_relation}, \texttt{multi\_hop\_path}, \texttt{conjunction\_intersection}, \texttt{shared\_intermediate}, \texttt{optional\_match}, \texttt{comparison}, \texttt{union\_disjunction}, \texttt{set\_equivalence}, \texttt{shortest\_path} \\
Return family & \texttt{name\_return}, \texttt{property\_return}, \texttt{distinct\_set}, \texttt{argmax\_topk}, \texttt{aggregation}, \texttt{existence\_boolean}, \texttt{comparison\_boolean}, \texttt{path\_length} \\
\midrule
\multicolumn{2}{@{}l}{\textbf{Pattern-conditioned match families $\mathcal{M}(q)$}} \\
$q=\langle h,\_,\_\rangle$ & \texttt{entity\_lookup}, \texttt{one\_hop\_relation} \\
$q=\langle h,r,*\rangle$ & \texttt{one\_hop\_relation}, \texttt{multi\_hop\_path}, \texttt{set\_equivalence} \\
$q=\langle h,*,t\rangle$ & \texttt{one\_hop\_relation}, \texttt{multi\_hop\_path}, \texttt{conjunction\_intersection}, \texttt{shared\_intermediate}, \texttt{comparison}, \texttt{shortest\_path} \\
$q=\langle h,r,t\rangle$ & \texttt{one\_hop\_relation}, \texttt{multi\_hop\_path}, \texttt{comparison}, \texttt{shortest\_path} \\
$q=\texttt{Hybrid}$ & \texttt{multi\_hop\_path}, \texttt{conjunction\_intersection}, \texttt{shared\_intermediate}, \texttt{optional\_match}, \texttt{comparison}, \texttt{union\_disjunction}, \texttt{set\_equivalence}, \texttt{shortest\_path} \\
\midrule
\multicolumn{2}{@{}l}{\textbf{Answer-conditioned return families $\mathcal{R}(a)$}} \\
$a=\texttt{entity}$ & \texttt{name\_return}, \texttt{property\_return}, \texttt{argmax\_topk} \\
$a=\texttt{set}$ & \texttt{name\_return}, \texttt{property\_return}, \texttt{distinct\_set} \\
$a=\texttt{number}$ & \texttt{property\_return}, \texttt{aggregation}, \texttt{path\_length} \\
$a=\texttt{boolean}$ & \texttt{property\_return}, \texttt{existence\_boolean}, \texttt{comparison\_boolean} \\
\midrule
\multicolumn{2}{@{}l}{\textbf{Difficulty-conditioned controls $(\mathcal{M}_{d},\mathcal{R}_{d})$}} \\
$d=\texttt{simple}$ & \textbf{Match:} \texttt{entity\_lookup}, \texttt{one\_hop\_relation}; \textbf{Return:} \texttt{name\_return}, \texttt{property\_return}, \texttt{distinct\_set}, \texttt{existence\_boolean}, \texttt{aggregation} \\
$d=\texttt{medium}$ & \textbf{Match:} \texttt{one\_hop\_relation}, \texttt{multi\_hop\_path}, \texttt{conjunction\_intersection}, \texttt{shared\_intermediate}, \texttt{comparison}; \textbf{Return:} \texttt{name\_return}, \texttt{property\_return}, \texttt{distinct\_set}, \texttt{argmax\_topk}, \texttt{aggregation}, \texttt{existence\_boolean}, \texttt{comparison\_boolean}, \texttt{path\_length} \\
$d=\texttt{hard}$ & \textbf{Match:} \texttt{multi\_hop\_path}, \texttt{conjunction\_intersection}, \texttt{shared\_intermediate}, \texttt{optional\_match}, \texttt{comparison}, \texttt{union\_disjunction}, \texttt{set\_equivalence}, \texttt{shortest\_path}; \textbf{Return:} \texttt{name\_return}, \texttt{property\_return}, \texttt{distinct\_set}, \texttt{argmax\_topk}, \texttt{aggregation}, \texttt{existence\_boolean}, \texttt{comparison\_boolean}, \texttt{path\_length} \\
\bottomrule
\end{tabular}
\caption{Quizzer generation controls and their compatibility constraints. The five controls specify answer form, abstract question structure, graph-access structure, return semantics, and intended complexity; all are soft generation targets rather than verified labels.}
\label{tab:quizzer-generation-controls}
\end{table*}

\textbf{Construction of a valid generation setting.} For $a\in\mathcal{A}$, $q\in\mathcal{Q}$, and $d\in\mathcal{D}$, a valid setting combines the sampled axes with families allowed by both relevant table constraints:
\begin{equation}
\begin{gathered}
g=(a,q,d,f_{\mathrm{match}},f_{\mathrm{return}}),\\
f_{\mathrm{match}}\in\mathcal{M}(q)\cap\mathcal{M}_{d},
\qquad
f_{\mathrm{return}}\in\mathcal{R}(a)\cap\mathcal{R}_{d}.
\end{gathered}
\label{eq:app-valid-generation-setting}
\end{equation}
This intersection-based construction prevents incompatible combinations, such as assigning a shortest-path objective to a simple entity-lookup question or requesting a Boolean return for a set-valued answer.

The sampled setting is supplied to the Quizzer as a soft generation target. The Quizzer may adapt its concrete exploration to the local graph schema and available seed neighborhood, but the generated candidate is not accepted solely because it follows the sampled controls. Its marked Cypher evidence is subsequently replayed and canonicalized, and the final question--answer pair is retained only if it passes execution certification and semantic curation.

\section{Data Backend, Prompts, and Tool Interfaces}
\label{app:prompt-tool-interfaces}

\subsection{Dataset Definition and Preprocessing}
\label{app:dataset-preprocessing}

We use the original GRBENCH graphs without changing their semantics: Academic is bibliographic, E-Commerce covers products and behavioral links, Literature covers books and publication metadata, Healthcare is biomedical, and Legal contains court and citation data. Each graph is imported into Neo4j while preserving identifiers, the typed directed schema, endpoint constraints, properties, and its domain-specific dataset label. We extract schema and property metadata and create indexes for Cypher access and textual entity retrieval; this preprocessing changes the access layer rather than graph facts or benchmark questions.

\subsection{Quizzer Protocol}
\label{app:quizzer-prompt-tools}

\textbf{Prompt construction.} The Quizzer system prompt combines a task-invariant interaction protocol with dynamically injected guidance for the sampled difficulty, question pattern, answer type, admissible match/return families, and domain terminology. The initial user message additionally provides a random seed node and its one-hop neighborhood, the available schema types, and the exploration-round budget. Thus, graph-specific entities and relations enter through runtime context rather than fixed few-shot examples.

\textbf{Quizzer interaction sequence.}
Interaction sequence of Quizzer is listed as follows:
\begin{enumerate}
\item \textbf{Initialize and explore.} Inject the sampled controls, anchor neighborhood, schema inventory, domain guidance, and round budget; each turn invokes exactly one \texttt{select\_schema} or \texttt{code} action.
\item \textbf{Retrieve schema and graph evidence.} The \texttt{select\_schema} tool returns typed outgoing patterns, relation meanings, and type-specific properties. The \texttt{code} tool provides persistent Python with synchronous read-only \texttt{cypher(query, params, limit)}: queries are typed, parameterized, and aliased; successful variables persist, whereas failed turns are rolled back.
\item \textbf{Fix the candidate.} 
Emit exactly one question, short reasoning sketch, and answer without internal graph identifiers; these fields are then held fixed unless the generation-time repair check below triggers a minimal QA rewrite.
\item \textbf{Construct evidence.} Recompute the answer in one final code block and call \texttt{mark\_evidence(...)} only on the bounded direct \texttt{TracedResult} objects used. The runtime checks trace identity, answer agreement, and complete marking of successful final-stage queries.
\item \textbf{Repair check.} After successful evidence execution, apply the constrained repair judge once. Minimally rewrite the QA, regenerate its evidence, and re-execute it if needed before certification.
\end{enumerate}

\noindent\textbf{Information isolation.}
\noindent Exploration probes are not automatically treated as gold evidence, and \texttt{mark\_evidence(...)} is available only after the candidate is fixed. The prompt exposes neither test-side templates or answers nor downstream verifier, semantic-Judge, Solver, or reward information.

\subsection{Solver Protocol}
\label{app:solver-prompt-tools}

\textbf{Prompt construction.} Each Solver rollout receives a domain-specific system prompt and the natural-language question as its sole user message. The system prompt supplies the dataset scope, typed directed schema, tool signatures, answer-format rules, and limited domain conventions. It is generated from schema metadata rather than task-local supervision and does not reveal the Quizzer's evidence, certified rubric, reference answer, verifier status, or reward.

\textbf{Two-level tool interface.} The Solver calls \texttt{node\_\allowbreak retriever} to map ambiguous entity mentions to candidate node IDs and types, and \texttt{code} to execute persistent sandboxed Python. Retriever output only anchors graph access and cannot be marked as evidence. Within \texttt{code}, synchronous read-only \texttt{cypher()} returns authenticated \texttt{TracedResult} objects, while \texttt{mark\_\allowbreak evidence(...)} selects the direct query results supporting the prediction.

\textbf{Solver interaction sequence.}
Interaction sequence of Solver is listed as follows:
\begin{enumerate}
\item \textbf{Resolve entities.} Retrieve each ambiguous mention once; identifiers explicitly supplied by the question are passed directly to Cypher.
\item \textbf{Query and compose.} Follow the directed typed schema with parameterized, explicitly aliased queries. One fixed graph operation remains in one query, whereas genuinely independent sets or operands are combined in persistent Python. Failed code calls roll back their variables and query artifacts.
\item \textbf{Select evidence.} After all supporting queries succeed, call \texttt{mark\_\allowbreak evidence(...)} on their direct \texttt{TracedResult} objects. The prompt requests one final marking; derived Python values and retriever outputs are invalid evidence.
\item \textbf{Submit the answer.} Evidence marking does not terminate the rollout. The Solver ends with a \texttt{\textbackslash answer\{\ldots\}} line, using complete entity names for set-valued questions, a bare number for counts, and \texttt{True}/\texttt{False} only for Boolean questions.
\end{enumerate}

\section{Certification and Semantic Curation}
\label{app:certification}

GraphCert treats each Quizzer output as a candidate certificate, not trusted supervision. A generated pair $(Q,A)$ and its marked support queries $\mathcal{C}_{\mathrm{gold}}$ must pass deterministic execution certification and rubric-guided semantic curation before entering $\mathcal{D}_{\mathrm{cert}}$.

\subsection{Certificate Format}
\label{app:certificate-format}

Each certificate is constructed in persistent Python through a read-only \texttt{cypher()} interface and \texttt{mark\_evidence()}. The support program executes one or more queries, composes a support answer, and marks the results used as evidence.

A successful query returns a list- or dictionary-like traced result with an immutable artifact identity. Its trace stores the Cypher text and parameters, execution status, row count, ordering and truncation metadata, normalized-result hash, and schema metadata. Because \texttt{mark\_evidence()} accepts only these traced results, an ordinary Python object cannot be presented as evidence.

The runtime synchronizes the artifact-ID, query, and evidence views, stored as \texttt{marked\_artifact\_ids}, \texttt{marked\_queries}, and \texttt{gold\_evidence}. Certification requires the same unique artifact identities in all three, rejecting missing objects, mismatched query--result pairs, and nonexistent executions.

The support program also recomputes an execution-grounded answer $A_{\mathrm{support}}$. After answer-type-specific normalization, certification requires
\begin{equation}
\operatorname{normalize}_{T}(A_{\mathrm{support}})
=
\operatorname{normalize}_{T}(A),
\label{eq:app-support-answer}
\end{equation}
where $T$ is the entity, Boolean, number, or set answer type. Thus, $A$ must be recoverable from the recorded interactions rather than merely plausible.

Each marked query retains at most 200 rows; the runtime requests one additional row and sets the truncation flag when that row exists. Truncated results remain valid partial observations but cannot certify complete sets, exact counts, set equality, or other full-result claims.

\subsection{Execution Certification Rules}
\label{app:execution-certification}

Execution certification consists of four complementary checks.

\textbf{Format, provenance, and safety.} The verifier requires $Q$, $A$, the support program, marked queries, and their evidence; it also checks unique, consistent artifact identities, successful executions, and generation-time hashes. Cypher must be read-only: \texttt{CREATE}, \texttt{MERGE}, \texttt{DELETE}, \texttt{SET}, \texttt{REMOVE}, \texttt{DROP}, \texttt{LOAD CSV}, \texttt{CALL}, and APOC procedures are rejected.

\textbf{Schema and static Cypher checks.} The verifier checks labels, relation types and directions, endpoints, properties, filters, returns, and active operations against $\mathcal{S}$. Direction validation uses the full source-label, relation-type, and target-label triple, so reversing a directed relation is invalid. The same analysis derives the evidence-reward representation: typed nodes and directed patterns, return expressions and contracts, result shape, deduplication, ordering, filtering, aggregation, counting, limiting, optional matching, and match/return families.

\textbf{Independent query replay.} Every marked query is independently re-executed against $\mathcal{G}$ with its recorded parameters. For replayed result $R_i^{\mathrm{replay}}$ and deterministic post-normalization hash $H(\cdot)$, certification requires
\begin{equation}
\begin{aligned}
H_i^{\mathrm{replay}}
&=H\left(
\operatorname{normalize}(R_i^{\mathrm{replay}})
\right)
=H_i^{\mathrm{trace}},\\
t_i^{\mathrm{replay}}&=t_i^{\mathrm{trace}}.
\end{aligned}
\label{eq:app-replay-consistency}
\end{equation}
where $t_i$ is the truncation flag. Explicitly ordered queries preserve row order; other results use unordered semantics. Execution errors, timeouts, hash or truncation mismatches, unsafe queries, and invalid provenance are hard failures, regardless of whether their source is model error, artifact corruption, parameter drift, unstable ordering, or a changed graph snapshot.

\textbf{Answer grounding and evidence completeness.} The verifier checks both that the support program produces $A$ and that the marked results suffice for the computation. Scalars, entities, Booleans, complete single-column sets, numeric aggregates, and deterministic row counts can be strictly grounded. A replayable, answer-consistent multi-query Python composition whose full lineage cannot be established statically is instead routed to judgeable partial certification.

\subsection{Program-Level Certification Outcomes}
\label{app:strict-partial-certificates}

Execution certification assigns each candidate one of three program-level statuses: judgeable strict, judgeable partial, or hard failure. The three outcomes distinguish complete static certification, bounded verifier abstention, and hard-gate rejection. Judgeable strict and judgeable partial may proceed to semantic curation, whereas hard failures are rejected before semantic review.

\textbf{Judgeable strict.} A certificate is judgeable strict when all static and replay checks pass, the execution-grounded answer matches $A$, and evidence completeness and direct grounding are established. This status measures verifiability, not difficulty: a reproducible multi-hop aggregate may be strict, while a simple query with an invalid direction, ambiguous return contract, or truncated result may not be.

\textbf{Judgeable partial.} A certificate is judgeable partial only when replay and every hard gate pass and the remaining uncertainty is allowlisted. Cases include complex Python composition; conservative ambiguity in conjunctions, shared variables, or generic relations; incomplete static support for distinctness or filters; uncertain return-contract derivation; and sampled-versus-realized family drift. Partial means verifier abstention, not acceptance: provenance, replayability, and basic grounding are confirmed, but some semantic dependency remains unproven.

Judgeable partial is not a lower-severity failure: it means that all non-negotiable execution, safety, provenance, and answer-grounding gates pass, while an allowlisted dependency remains unproven statically. Hard failure instead denotes violation of at least one such gate.

\textbf{Hard failure.} Hard failure covers invalid structure or artifact membership, unsafe or schema-invalid queries, execution or replay failure, answer--evidence inconsistency, missing evidence, exact-set or exact-count claims from truncated results, and unresolved issues outside the allowlist. These candidates do not reach the semantic Judge.

Thus, static or replay failures are hard failures; fully grounded candidates are judgeable strict; and replayable candidates with only allowlisted uncertainty are judgeable partial. Neither judgeable status enters $\mathcal{D}_{\mathrm{cert}}$ without semantic curation.

\subsection{Semantic Curation Protocol}
\label{app:semantic-curation}

Replay establishes executability, not complete support for the natural-language meaning of $(Q,A)$. GraphCert therefore uses the Quizzer's local model for a two-stage rubric-guided semantic review. This semantic Judge is separate from the blinded Judge used only in post-hoc QA-quality analysis; the admission pipeline uses no human review.

The Judge receives $Q$, $A$, the support program, bounded query/evidence previews, and canonical per-query derivations. It does not receive the full exploration transcript, free-form reasoning, Solver trajectory, or sampled family, reducing bias from the generation request or a plausible-looking trace.

\textbf{Stage 1: evidence-sufficiency reflection.} Stage 1 requires all certificate objects and a \texttt{mark\_evidence()} call, then checks:

\begin{itemize}
    \item grounding of entities, constraints, and descriptive modifiers;
    \item correct relation types, directions, paths, and verbalization;
    \item correct quantifiers and cross-branch identity constraints;
    \item correct answer type, scope, deduplication, completeness, and counted
    object; and
    \item actual use of graph results rather than hard-coded answers.
\end{itemize}

Semantic failure causes rejection; infrastructure failures such as timeouts are recorded as uncertain rather than data-quality failures.

\textbf{Stage 2: focused audit.} Stage 2 audits claim coverage, specificity, modifiers, relation verbalization, quantifier scope, entity grounding, path faithfulness, answer scope, and unsupported claims. Any required-check failure prevents acceptance.

The final semantic verdict is one of \texttt{accept}, \texttt{reject}, or \texttt{uncertain}. Only \texttt{accept} is admitted into the certified training set:
\begin{equation}
\begin{aligned}
\mathcal{D}_{\mathrm{cert}}
=\{x:\;&
s_{\mathrm{program}}(x)
\in\{\text{strict},\text{partial}\},\\
&
j_{\mathrm{semantic}}(x)=\text{accept}\}.
\end{aligned}
\label{eq:app-certified-set}
\end{equation}
The \texttt{uncertain} verdict is excluded from both admission and the reported semantic-rejection count, so infrastructure errors are not labeled as quality failures. An accepted strict certificate is program-strict and semantic-accepted; an accepted partial certificate is program-partial and semantic-accepted. Neither status claims an error-free Judge, formally proven causal lineage, unique query solution, or invariance to future graph snapshots.

\section{Evidence-Score Component Definitions}
\label{app:evidence-scoring}

The main text introduces query alignment, aggregate coverage, and the answer-primary reward. Here we provide the complete component definitions and fixed implementation settings needed to reproduce these scores.

\textbf{Multiplicity-aware feature F1.}
For canonical feature multisets $X$ and $Y$, let $o(X,Y)$ be the sum, over all features, of the smaller multiplicity in the two multisets. We compute
\begin{equation}
\operatorname{MF1}(X,Y)=\frac{2o(X,Y)}{|X|+|Y|}.
\label{eq:app-mf1}
\end{equation}
Repeated features therefore retain their multiplicity, while deterministic serialization makes the score invariant to ordering. We set $\operatorname{MF1}(\emptyset,\emptyset)=1$ and return zero when exactly one multiset is empty.

\textbf{Pair-score components.}
For a query $q$, the deriver extracts typed and directed graph patterns $\mathcal{P}_q$, participating nodes $\mathcal{V}_q$, returned expressions $\mathcal{R}_q$, and active operations $\mathcal{O}_q$. It also records result shape $h_q$, returned-property signature $p_q$, deduplication flag $d_q$, ordering flag $r_q$, and match/return families $f_q^{\mathrm{match}}$ and $f_q^{\mathrm{return}}$. The structural components of the pair score are
\begin{equation}
\begin{aligned}
S_{\mathrm{shape}}(i,j)={}&0.80\operatorname{MF1}(\mathcal{P}_i,\mathcal{P}_j)
+0.20\operatorname{MF1}(\mathcal{V}_i,\mathcal{V}_j),\\
S_{\mathrm{ops}}(i,j)={}&\operatorname{MF1}(\mathcal{O}_i,\mathcal{O}_j),\\
S_{\mathrm{family}}(i,j)={}&
0.50\mathbb{1}[f_j^{\mathrm{match}}\ne\emptyset]
\mathbb{1}[f_i^{\mathrm{match}}=f_j^{\mathrm{match}}]\\
&+0.50\mathbb{1}[f_j^{\mathrm{return}}\ne\emptyset]
\mathbb{1}[f_i^{\mathrm{return}}=f_j^{\mathrm{return}}].
\end{aligned}
\label{eq:app-structural-components}
\end{equation}
The return-contract component is
\begin{equation}
\begin{aligned}
S_{\mathrm{contract}}(i,j)={}&
0.45\operatorname{MF1}(\mathcal{R}_i,\mathcal{R}_j)
+0.20\mathbb{1}[h_i=h_j]\\
&+0.15\mathbb{1}[p_i=p_j]
+0.10\mathbb{1}[d_i=d_j]\\
&+0.10\mathbb{1}[r_i=r_j].
\end{aligned}
\label{eq:app-contract-component}
\end{equation}
Result shape distinguishes scalar, row, and set outputs. Active operations include filtering, aggregation, counting, ordering, limiting, distinctness, and optional matching.

The complete pair score assigns weights $0.55$ to graph shape, $0.25$ to the return contract, $0.15$ to active operations, and $0.05$ to match/return families. Consequently, directed patterns and participating nodes account for $0.44$ and $0.11$ of the complete pair score. All components lie in $[0,1]$; a failed or empty static derivation sets the complete pair score to zero.

\textbf{Aggregate-score components.}
For aggregate coverage, the deriver pools features across all valid queries on each side. Let $\operatorname{Rel}(\mathcal{C})$ denote the set of referenced relation types, $\operatorname{Pat}(\mathcal{C})$ the multiset of canonical directed patterns, $\operatorname{Node}(\mathcal{C})$ the multiset of participating nodes, $\operatorname{Prop}(\mathcal{C})$ the set of referenced properties, and $\operatorname{Ret}(\mathcal{C})$ the multiset of per-query return requirements.

We first compute pooled node and property coverage:
\begin{equation}
\begin{aligned}
S_{\mathrm{nodeprop}}={}&
0.60\operatorname{MF1}\left(
\operatorname{Node}(\mathcal{C}_{\mathrm{pred}}),
\operatorname{Node}(\mathcal{C}_{\mathrm{gold}})
\right)\\
&+0.40\operatorname{MF1}\left(
\operatorname{Prop}(\mathcal{C}_{\mathrm{pred}}),
\operatorname{Prop}(\mathcal{C}_{\mathrm{gold}})
\right).
\end{aligned}
\label{eq:app-nodeprop}
\end{equation}

The aggregate structural-coverage score is
\begin{equation}
\begin{aligned}
S_{\mathrm{aggregate}}={}&
0.25\operatorname{MF1}\left(
\operatorname{Rel}(\mathcal{C}_{\mathrm{pred}}),
\operatorname{Rel}(\mathcal{C}_{\mathrm{gold}})
\right)\\
&+0.35\operatorname{MF1}\left(
\operatorname{Pat}(\mathcal{C}_{\mathrm{pred}}),
\operatorname{Pat}(\mathcal{C}_{\mathrm{gold}})
\right)\\
&+0.20S_{\mathrm{nodeprop}}\\
&+0.20\operatorname{MF1}\left(
\operatorname{Ret}(\mathcal{C}_{\mathrm{pred}}),
\operatorname{Ret}(\mathcal{C}_{\mathrm{gold}})
\right).
\end{aligned}
\label{eq:app-aggregate-components}
\end{equation}

Each return requirement contains the canonical returned expressions, returned-property signature, deduplication flag, and result shape. Failed derivations are excluded from the pooled features. If either side has no valid derivation, then $S_{\mathrm{aggregate}}=0$.

\textbf{Implementation settings.}
The evidence mixture uses $\eta=0.70$, and the answer-primary reward uses $\lambda=0.10$. Query alignment uses maximum-weight one-to-one matching between the marked query sets. Only explicitly marked queries are scored. If either marked query set is empty, the evidence score is zero; if no final answer can be parsed, the rollout reward is zero.

\section{Experimental Details and Additional Analyses}
\label{app:additional-analyses}

\subsection{GRPO Hyperparameters}
\label{app:grpo-hyperparameters}

We implement GRPO with \texttt{verl}. Each domain-specific model is trained on 2,000 certified examples for 400 steps, with eight Solver trajectories sampled per prompt at temperature $1.0$. Table~\ref{tab:grpo-hyperparameters} reports the remaining settings; training uses gradient checkpointing, padding removal, sample packing, dynamic batching, and Flash Attention. The evidence mixture weight $\eta$ and reward weight $\lambda$ are 0.70 and 0.10.

\begin{table}[ht]
\centering
\small
\begin{tabular}{ll}
\toprule
Hyperparameter & Value \\
\midrule
RL algorithm & GRPO \\
Training framework & \texttt{verl} \\
Training samples per domain & 2,000 \\
Optimization steps & 400 \\
Responses per prompt & 8 \\
Rollout temperature & 1.0 \\
Actor learning rate & $1\times10^{-6}$ \\
Clipping coefficient $\epsilon$ & $0.20$ \\
Initial KL coefficient & $1\times10^{-4}$ \\
Training batch size & 16 \\
Rollout batch size & 16 \\
Micro training batch size & 8 \\
Micro rollout batch size & 16 \\
Maximum prompt length & 4,096 \\
Maximum response length & 8,192 \\
Numerical precision & \texttt{bfloat16} \\
\bottomrule
\end{tabular}
\caption{GRPO hyperparameters used for GraphCert.}
\label{tab:grpo-hyperparameters}
\end{table}

\subsection{Computational Resources}
\label{app:computational-resources}

The 4B and 8B experiments use the separate Linux-server configurations in Table~\ref{tab:hardware-configuration}. Unless otherwise specified, domain-specific training and evaluation use the 4B configuration; the 8B server is used only for model-scaling experiments.

\begin{table}[ht]
\centering
\small
\begin{tabular}{@{}
>{\raggedright\arraybackslash}p{0.27\columnwidth}
>{\raggedright\arraybackslash}p{0.29\columnwidth}
>{\raggedright\arraybackslash}p{0.33\columnwidth}
@{}}
\toprule
Configuration & 4B experiments & 8B experiments \\
\midrule
CPU & Intel Xeon Gold 6342, 2.80\,GHz &
Intel Xeon Platinum 8358P, 2.60\,GHz \\
\midrule
GPU &
$4\times$ NVIDIA A40 &
$4\times$ NVIDIA A800-SXM4-80GB \\
System memory & 1\,TB & 1\,TB \\
CUDA & 12.9 & 12.3 \\
Operating system &
Ubuntu 22.04.5 LTS &
Ubuntu 22.04.3 LTS \\
\bottomrule
\end{tabular}
\caption{Hardware and software configurations used for the 4B and 8B experiments.}
\label{tab:hardware-configuration}
\end{table}

\subsection{Baseline Details}
\label{app:baseline-details}

We compare methods spanning parametric-only answering, passive retrieval, symbolic querying, iterative graph interaction, and graph-grounded post-training. Their assumptions and graph interaction mechanisms are detailed below.

\textbf{BaseLLM.} This non-graph baseline measures how much domain-specific knowledge is already encoded in the backbone. The model receives only the question: it has no access to the graph, retrieved context, or graph tools. Its result therefore separates gains from explicit graph grounding from those attributable to parametric knowledge alone.

\textbf{TextRAG.} TextRAG~\cite{ragsurveytextrag} treats the graph as a text corpus. Node attributes and edge facts are linearized into text chunks, a dense retriever selects chunks by semantic similarity to the question, and the retrieved text is appended to the answering prompt. Retrieval is passive and does not preserve an executable graph structure; the model reasons over the selected textual context.

\textbf{GraphRAG.} GraphRAG~\cite{ye-etal-2024-language-graphrag} adds local topology to retrieval. It identifies core entities mentioned in the question, expands their two-hop neighborhoods, and linearizes the resulting subgraphs for the LLM. Compared with TextRAG, this exposes connected relational context rather than isolated text chunks, but graph access remains a one-shot retrieval stage followed by answer generation.

\textbf{Cypher.} The Cypher baseline uses the LLM to produce an executable query over the underlying graph. A node retriever first grounds question mentions to candidate nodes; conditioned on these candidates and the question, the model generates Cypher, executes it, and receives the returned nodes, attributes, or aggregates for final answer synthesis. This provides symbolic graph access, but places the burden of expressing the required reasoning in the generated query.

\textbf{GraphCoT.} GraphCoT~\cite{jin2024graphcot} is an iterative reasoning baseline that interleaves chain-of-thought reasoning with calls to a predefined graph-traversal interface. At each step, the model selects a graph operation, observes the retrieved evidence, and updates its reasoning context before choosing the next action. This incremental interaction can avoid the information loss of a single long linearized subgraph and supports multi-step evidence collection.

\textbf{PolyG.} PolyG~\cite{liu2025polyg} performs adaptive graph traversal rather than committing to a fixed retrieval radius. After candidate entities are grounded by node retrieval, the model selects graph operators and exploration paths according to the question and its current observations. The traversal policy can therefore vary across reasoning steps and question types, including cases that require non-uniform paths or intermediate constraints.

\textbf{GraphCounselor.} GraphCounselor~\cite{gao2025graphCounselor} organizes graph reasoning as collaboration among agents responsible for planning, graph execution, and reflection. Execution results are returned to the group, and reflection can revise the plan, correct earlier decisions, or adjust the remaining reasoning depth. It thus represents a multi-agent alternative to the single-agent iterative methods above.

\textbf{GraphScout.} GraphScout~\cite{graphscout2026} is the closest post-training baseline: it uses a strong Quizzer to explore the graph and synthesize question--answer supervision, then trains a Solver with answer-based reward. For GraphScout(4B), we replace its strong proprietary Quizzer with Qwen3-4B-Instruct-2507 and initialize the Solver from the same checkpoint, while retaining GraphScout's data-generation and answer-reward training procedure. This controlled variant isolates the effect of learning from a compact self-Quizzer. The GPT-4o~\cite{achiam2023gpt4} and DeepSeek-V3.2~\cite{2025deepseekv32} prompting rows in the main results, together with the strong-Quizzer GraphScout comparison, are taken from the published GraphScout evaluation rather than rerun.

\subsection{Token Efficiency}
\label{app:token-efficiency}

\begin{figure}[htbp]
\centering
\includegraphics[width=\columnwidth]{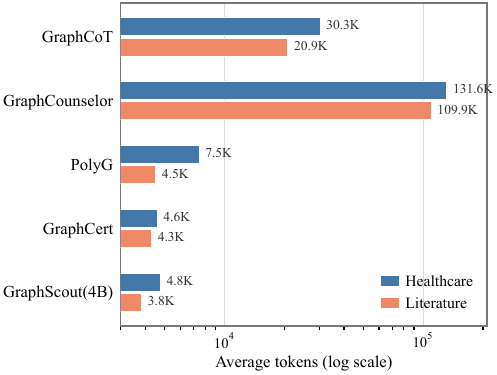}
\caption{Average token consumption on Healthcare and Literature; lower is better. The horizontal axis uses a logarithmic scale.}
\label{fig:token-consumption}
\end{figure}

Figure~\ref{fig:token-consumption} compares average token consumption on Healthcare and Literature. GraphCert remains in the same low-token regime as GraphScout(4B), while using substantially fewer tokens than the other active graph-reasoning baselines; its performance therefore does not rely on longer inference traces.

\subsection{Performance by Question Difficulty}
\label{app:difficulty-analysis}

\begin{figure}[htbp]
\centering
\includegraphics[width=\columnwidth]{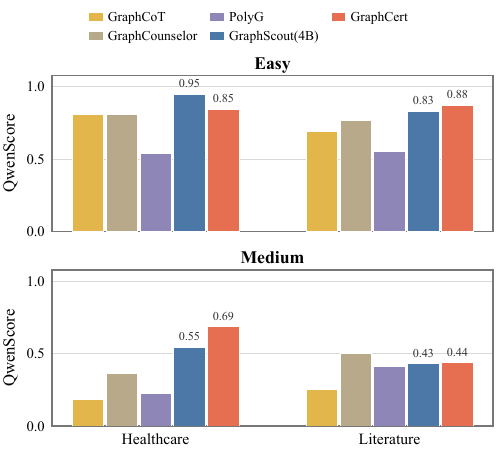}
\caption{QwenScore by question difficulty on the informative Easy and Medium splits of Healthcare and Literature.}
\label{fig:difficulty-performance}
\end{figure}

Figure~\ref{fig:difficulty-performance} presents a difficulty-stratified analysis and reports QwenScore on the informative Easy and Medium splits. GraphCert leads on Easy Literature and Medium Healthcare, and is comparable to GraphScout(4B) on Medium Literature. Hard is not plotted because Healthcare contains no Hard questions. Literature is built from Goodreads, where many Hard questions emphasize recommendation-style reasoning and external-world knowledge rather than structured graph traversal; this makes the split less diagnostic of graph exploration, and all compared methods score zero on it.

The Easy Healthcare gap is largely associated with the reversed-relation cases examined in the sensitivity analysis below. Although these questions are easy in traversal depth, their wording requests the inverse of a directed schema relation. GraphScout(4B) more often follows that reversed wording, which can match the benchmark reference, whereas GraphCert tends to preserve the stored schema direction. Thus, the lower Easy score partly reflects stricter schema faithfulness rather than a need for deeper exploration.

\subsection{Accepted Partial Certificates and Coverage}
\label{app:partial-distribution}

We compare semantically accepted strict and partial certificates from one biomedical run. Both passed the same semantic gate; partial means that the static verifier abstained on an allowlisted dependency despite successful replay and answer reproduction. Families were independently derived from the QA, support program, marked evidence, and normalized queries. Adding partials more than doubled pair coverage, from 18 to 39, chiefly recovering comparison, set-equivalence, conjunction, shared-intermediate, Boolean, and multi-query problems rare in the one-hop-heavy strict pool.

\begin{figure}[H]
\centering
\includegraphics[width=\columnwidth]{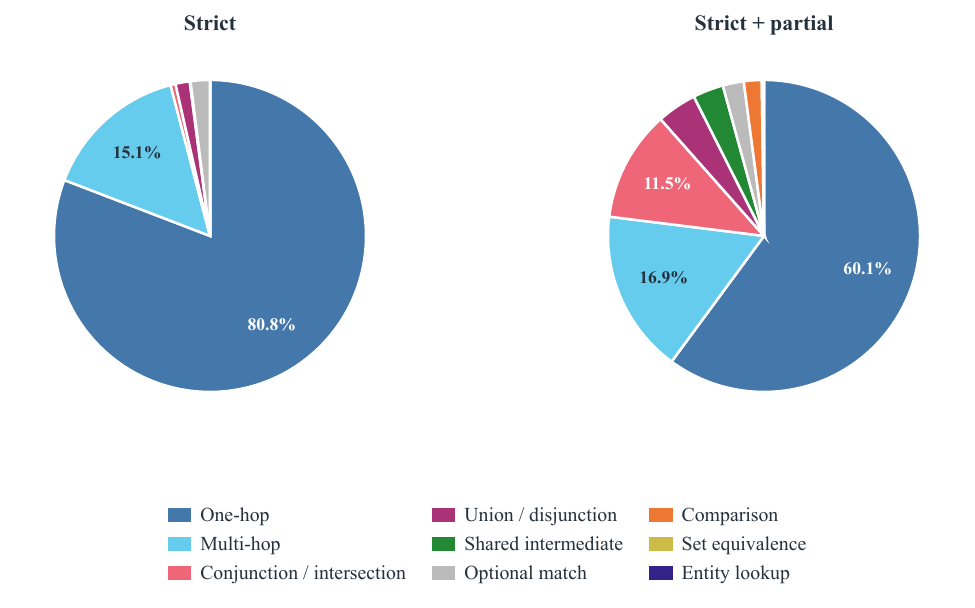}
\caption{Match-family distributions of semantically accepted questions in the biomedical run, comparing strict certificates with the combined strict-and-partial pool.}
\label{fig:strict-partial-match-family}
\end{figure}

\begin{figure}[H]
\centering
\includegraphics[width=\columnwidth]{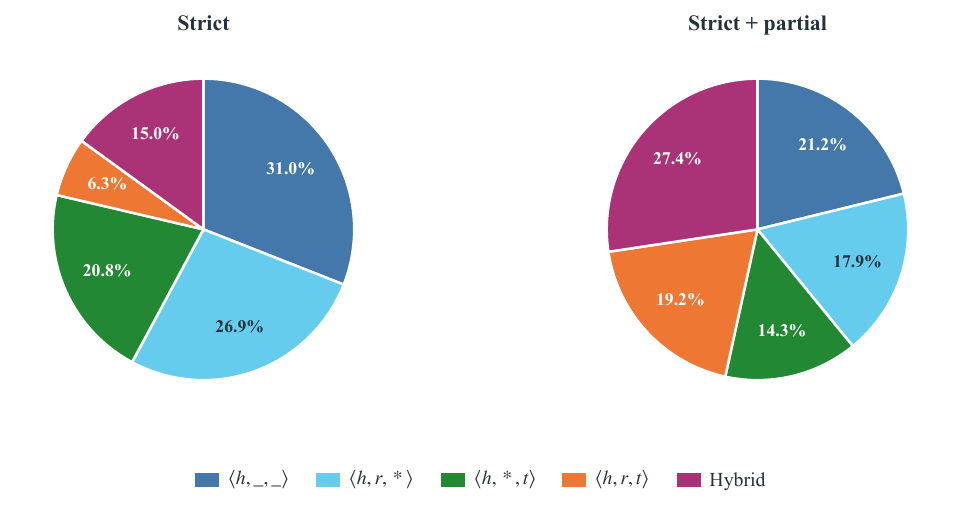}
\caption{Question-pattern distributions of semantically accepted questions in the biomedical run, comparing strict certificates with the combined strict-and-partial pool.}
\label{fig:strict-partial-question-pattern}
\end{figure}

Figures~\ref{fig:strict-partial-match-family} and~\ref{fig:strict-partial-question-pattern} show the corresponding match-family and question-pattern distributions: adding accepted partial certificates reduces one-hop dominance and broadens structural coverage, including hybrid questions. The gain persisted under sample-size matching and a target-aligned-only comparison, so it is not solely a volume or target-drift effect. This run-specific analysis measures coverage rather than strict proof: retained partials passed replay and semantic curation, but their Python-composition lineage was not statically proven.

\subsection{Healthcare Benchmark Sensitivity}
\label{app:healthcare-sensitivity}

A schema-grounded review flagged 56 of 270 Healthcare questions: most invert a directed relation, while the remainder equate symptom sets where the reference uses containment. We retain them in the official result. Excluding them changes GraphCert from 0.744/0.722 to 0.835/0.802 QwenScore/F1 on 214 questions, but we do not use this score for ranking because the baselines were not reevaluated.

\end{document}